\documentclass[runningheads]{llncs}

\usepackage[mobile]{eccv}

\usepackage{eccvabbrv}

\usepackage{mathtools}
\usepackage{amsmath,amssymb}
\usepackage{graphicx}
\usepackage{booktabs}
\usepackage{subcaption}
\usepackage{tabularx}
\usepackage{multicol}
\usepackage{multirow}

\usepackage[accsupp]{axessibility}  

\usepackage{hyperref}

\usepackage{orcidlink}

\DeclareMathOperator*{\argmin}{arg\,min}

\makeatletter

\catcode`@=11
\def\diamondop#1#2{%
  \mathbin{%
    \vcenter{%
      \hbox{%
        \setbox\z@=\hbox{$\m@th#1\oplus$}%
        \dimen@=\ht\z@ \advance\dimen@ \dp\z@
        \dimen@=1.1\dimen@
        \ooalign{%
          \resizebox{!}{\dimen@}{%
            \rotatebox[origin=c]{45}{$\m@th#1\square$}%
          }\cr
          \hfil$\m@th#1#2$\hfil\cr
        }%
      }%
    }%
  }%
}

\def\diamondplus{\mathpalette\diamondplus@{}}
\def\diamondplus@#1#2{\diamondop{#1}{+}}

\def\diamondminus{\mathpalette\diamondminus@{}}
\def\diamondminus@#1#2{\diamondop{#1}{-}}
\catcode`@=12

\makeatother

\newcommand{\presup}[2]{\prescript{#1}{}{#2}}

\definecolor{tabblue}{RGB}{31,119,180}   
\definecolor{taborange}{RGB}{255,127,14} 
\definecolor{tabgreen}{RGB}{44,160,44}   

\begin{document}

\title{PixVOD: Pixel-Distributed Direct Visual Odometry and Depth Estimation} 

\titlerunning{PixVOD}

\author{Shinjeong Kim\orcidlink{0000-0002-3199-4111} \and
Ignacio Alzugaray\orcidlink{0000-0002-7121-0000} \and
Callum Rhodes\orcidlink{} \and
Paul H. J. Kelly\orcidlink{0000-0001-5905-1804} \and
Andrew J. Davison\orcidlink{0000-0002-3784-099X}}

\authorrunning{S.~Kim et al.}

\institute{Department of Computing, Imperial College London\\
\email{\{s.kim, i.alzugaray, c.rhodes, p.kelly, a.davison\}@imperial.ac.uk}}

\maketitle

\begin{abstract}
Images composed of 2D pixel arrays are the standard input to computer vision algorithms, yet many underlying computations can be distributed across pixels. Transmitting raw, redundant, and noisy pixel data off the sensor remains inefficient, motivating a shift toward focal-plane sensor-processors that perform a significant part of the computation directly within each pixel. We envision pixels synthesizing higher-level signals locally, reducing downstream load, and providing richer inputs for higher-level vision tasks.

We propose a fully parallelizable form of visual odometry and depth estimation across pixels, where sensor-processors exchange information through Gaussian Belief Propagation (GBP) to achieve consensus about camera motion and infer depth from per-pixel photometric observations and a surface normal prior. 
To maintain geometric stability during optimization, we introduce a keyframe-like anchoring mechanism that regulates the effective baseline between frames, enabling consistent motion and depth updates.
Our method is evaluated on realistic datasets, demonstrating the feasibility of GBP-based pixel-level distributed odometry and depth estimation with keyframe anchoring on-sensor. Project Page: \url{https://www.shinjeongkim.com/pixvod/}
\end{abstract}

\begin{figure}
    \centering
    \includegraphics[width=0.75\linewidth]{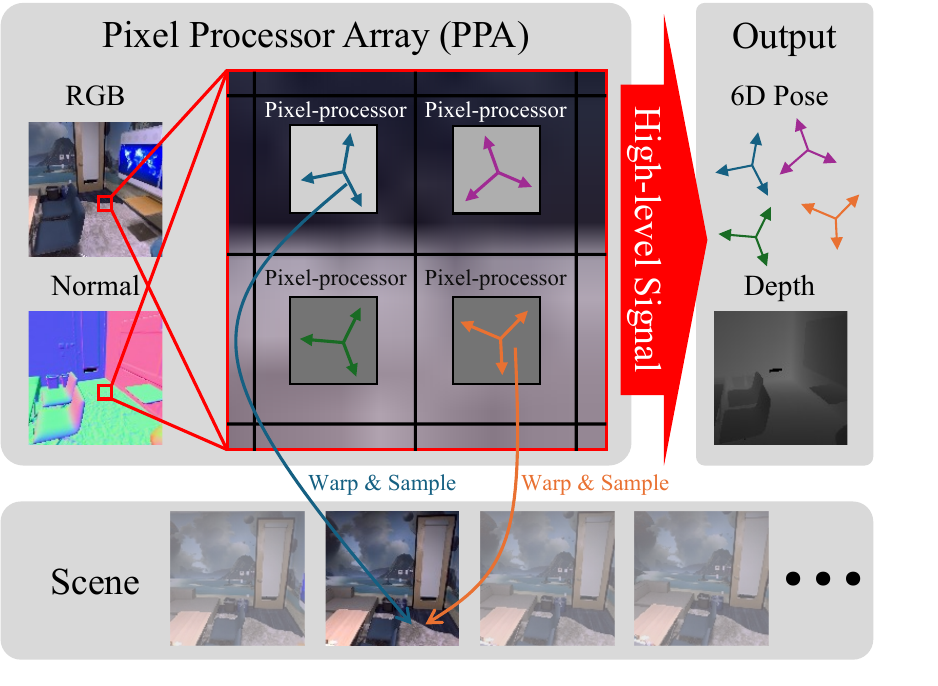}
    \vspace{-0.35cm}
    \caption{Most vision algorithms assume the use of an off-sensor central processor with access to all pixel information. In contrast, we present an algorithm to track camera pose and estimate depth on the assumption that each pixel has its own local processing capability, with only local memory and the ability to exchange messages with other pixels. Our goal is to motivate and guide the design of near-future chips able to carry out fundamental early vision tasks on-sensor, with extremely high efficiency and speed.}
    \label{fig:teaser}
    \vspace{-0.65cm}
\end{figure}

\section{Introduction}
\label{sec:intro}
Mobile edge applications like robotics or wearables demand results from computer vision at high frame rates, whilst maintaining low latency, high accuracy, and especially with {\em minimal energy consumption} in battery-driven products. A promising and fundamental path to achieving all of these goals is to move away from the standard separation of camera and processor, connected by video transmission, towards hardware which combines visual sensing and processing in a single unit. This could process visual input locally, rapidly, without high bandwidth data flow, and output only the necessary high-level estimates for the application in question. 

At the limit, in the {\em On Sensor Vision} paradigm, processing can
be built into a camera's image sensor itself. Prototypes already
exist, such as the SCAMP series of vision chips \cite{carey2013100, bose2023live}.  These implement a Pixel Processor Array (PPA) design, where each photo-sensitive pixel in a standard rectangular grid has a programmable local processing capability, a set of memory registers, and the ability to exchange information with neighbouring pixels. Crucially, a PPA achieves maximum performance when it uses {\em only} this local, pixelwise memory.

Despite the limited computational capacity of these chips, impressive image processing capabilities~\cite{bose2025descriptor,murai2020bit,so2024pixelrnn} have already been demonstrated, such as feature extraction and matching, local motion estimation, object tracking, and image classification and segmentation via a small CNN, with extremely high speed and low power usage.  Research is now underway to design the next generation of such chips, with questions about how much pixel-wise computation and memory, and what communication structure will enable more capability.

In this paper, we focus on the fundamental vision competence of local
6DoF motion estimation from a single camera, often known as visual
odometry (VO), and investigate how this could be achieved with
pixel-parallel computation to be suitable for a near-future on-sensor
vision device.  Previous work harnessing SCAMP for VO and SLAM has
performed feature extraction and even matching \cite{bose2025descriptor,murai2020bit}
on-sensor, but these methods have required frame-rate communication with an external
CPU to solve the optimisation needed for 6DoF motion estimation.

Here, we propose an algorithm to achieve 6DoF VO estimation with pure in-pixel computation.
The key challenge here is how to achieve consistent estimates of
camera motion, a global property, across the distributed memory
structure of a PPA.
We argue that a {\em dense} VO approach, where camera motion is estimated alongside a full dense depth map, is actually easier to achieve on a PPA than a sparse approach where geometry is estimated only for sparse points. This is because of the continuity of real scenes, which can be harnessed as priors on dense depth and therefore can be implemented as local pixel-wise constraints.

We formulate our approach as a distributed factor graph, stored across the whole pixel array with variables at every pixel to estimate camera motion and scene depth. We use identity factors to synchronize a global camera motion estimate across the whole array. 

\section{Related Work}


There is a growing amount of work exploring vision algorithms parallelized for processors with local computation and memory.
The SCAMP PPA vision chips have been used to enable high-speed and robust 
keypoint tracking and matching
\cite{murai2020bit, bose2025descriptor}, and this has been proven as the front-end to visual odometry, but with the use of an external processor for 3D estimation.
Meanwhile, the same chips have also been shown to be capable of running small CNN and RNN networks for classification and action recognition
\cite{Liu:etal:Frontiers2022,so2024pixelrnn}.

Other papers which are closest to our work have started to explore whether global geometric estimation can be achieved fully in-pixel. This work, like ours, was not yet implemented on true PPA chips due to the current limitations of hardware like SCAMP, which has a tiny amount of analogue and digital memory per pixels, can only perform simple arithmetic computations, and has a strict grid-wise pixel communication structure. Instead, it has aimed to explore the design space for near future chips with extra capabilities.
Scona \emph{et al.} \cite{scona2022scene} investigated 6DoF visual odometry for RGB-D sensors, with a model of pixel-parallel processing with both local and long-distance random communication structure implemented using a Graphcore 
IPU~\cite{ltdIPUProcessors}, while 
PixRO \cite{alzugaray2024pixro} studied 
in-pixel camera rotation tracking and compared the value of gridwise and hierarchical communication structured.
Both of these papers adopted factor graph and GBP formulations similar to our own work, where we now tackle the full monocular 6DoF visual odometry and depth estimation problem.



As Structure from Motion (SfM)~\cite{liu20233d, schonberger2016structure, agarwal2011building} has been practically utilized as an annotation method for 3D datasets, there have also been efforts to parallelize or accelerate its camera pose optimization.
Theoretically-oriented approaches, such as rotation~\cite{chatterjee2013efficient} and translation~\cite{zhuang2018baseline, pan2024global} averaging, achieve faster convergence by carefully analyzing, reformulating, and/or relaxing the underlying optimization constraints.
On the implementation side,~\cite{li2025fastmap, ortiz2020bundle} aims to accelerate computation by leveraging a deeper understanding of actual hardware architectures.
However, these methods focus on optimizing camera poses at the image level, rather than pursuing pixel-level parallelization.

In Visual SLAM, there have been several recent investigations of accelerations with specialized processing hardware~\cite{fang2017fpga, boikos2016semi, qin2019pi, asgari2020pisces, vemulapati2022fslam}. Our focus on camera tracking and depth estimation is aligned with speeding up the tracking-and-mapping frontend~\cite{lentaris2015hw, mandal2019visual, zhang2017visual, nikolic2014synchronized}. Nevertheless, most efforts in this vein parallelize only resource-intensive subroutines on specialized hardware, and the overall pipeline remains centralized.

It is well established that monocular depth estimation is highly dependent on the global context, modelled with globally connected Markov Random Fields~\cite{yuan2022neural} or Transformers~\cite{bochkovskiy2025depthpro, wang2025moge}.
In contrast, surface normal estimation has been shown to achieve remarkable performance even with models that possess only local receptive fields~\cite{bae2024rethinking}, empirically demonstrating that surface normals can be effectively estimated from relatively local information compared to depth~\cite{qi2018geonet, qi2020geonet++}.
This observation justifies our assumption of a per-frame normal prediction being available from a PPA.


Normal integration~\cite{cao2021normal, heep2024adaptive, kim2024discontinuity, cao2022bilateral} is the problem of reconstructing depth from a surface normal field, and in camera pose tracking and mapping contexts, numerous methods have explored its use to reconstruct depth or enhance its quality, or to be jointly optimized with other geometric objectives~\cite{pataki2025mp, zhu2024nicer, mazur2024superprimitive}.
However, these approaches typically assume a conventional centralized off-sensor processor.
In our work, we exploit the spatial sparsity inherent in the optimization formulation of the recent normal integration method BINI~\cite{cao2022bilateral} to enable efficient pixel-level parallelization.

Probabilistic graphical models~\cite{bishop2006pattern} and their associated optimization techniques~\cite{peterson1989explorations, zhang2002mean, pearl2014probabilistic} have been widely used long before the rise of deep learning. Even after deep learning–based approaches have become dominant, they have remained prevalent as post-processing tools for fine-grained refinement~\cite{krahenbuhl2011efficient}, as optimization back-ends for learning~\cite{yi2021differentiable, nabarro2023learning, guan2024neural}, and as powerful inference mechanisms in resource-constrained settings~\cite{nagata2023tangentially}. In our work, we model the camera pose and depth estimation problem as a Gaussian factor graph~\cite{dellaert2017factor} and employ Gaussian Belief Propagation to perform pixel-level distributed inference. The GBP formulation is naturally amenable to distributed optimization in many types of graph structure~\cite{ortiz2020bundle,murai2023robot}.

\section{Methodology}

\subsection{Direct Visual Tracking with Centralized Computation}\label{sec:centralized}

Before explaining the full details of our method, for clarity and as a baseline for later comparison, we first explain the simpler standard formulation of direct visual 6DoF
tracking against a rigid scene when computation and memory are centralized.

We consider a calibrated camera that undergoes general 6DoF motion in a static scene. Given  $\mathcal{I}_s$ (the ``source image'', from time 0) and $\mathcal{I}_t$ (the ``target image'', from time 1), we seek the relative pose $\mu_{[0:6]} \in \mathbb{SE}(3)$ and a source-view depth map $\exp{(\mu_{[6]})}\in D\in\mathbb{R}_+^{H\times W}$ that jointly explain their photometric differences.

In high frame-rate tracking, inter-frame camera motion and visual changes are small and we can perform direct, iterative photometric alignment. Formally, we align the warped target image $\mathcal{I}_t$ and the source image $\mathcal{I}_s$, over the overlapping domain $\Omega(\mu)$ induced by the camera motion parameterized by $\mu$. The resulting nonlinear problem is:
\begin{equation}
\mu^* = \underset{\mu}{\argmin}\sum_{p\in\Omega(\mu)}{\rho\left(\left\lVert\mathcal{I}_s\left[p\right]-\mathcal{I}_t\left[\mathcal{W}(p;\mu)\right]\right\rVert\right)^2}, \label{eq:photo}
\end{equation}where $\rho$ denotes a robust cost function designed to effectively ignore pixels whose photometric residuals are unreliable due to large correspondence errors, thereby enhancing the robustness of the system.

As in~\cite{newcombe2011dtam, scona2022scene}, the warping function $\mathcal{W}$, geometrically maps each pixel location $p$ in $\mathcal{I}_s$ and its corresponding depth $\exp{(\mu_{[6]})}$ to the corresponding location in $\mathcal{I}_t$:
\begin{equation}
\mathcal{W}(p; \mu) = \pi\left(K\exp{(\mu_{[0:6]})}K^{-1}\pi^{-1}(p, \exp{(\mu_{[6]})})\right)~,\label{eq:warp}
\end{equation} 
where the projection function is defined as $\pi(\mathbf{x}) = [x/z, y/z]^T$, and $K$ is the camera intrinsic matrix. Note that here, the left multiplication of $\exp{(\mu_{[0:6]})}$ is considered as an application of rigid transformation.

When optimizing Eq.~\eqref{eq:photo}, if all pixel information is accessible at each iteration, an optimal update can be computed for the current linearized model. However, this assumes that the entire image is loaded into globally accessible memory, which corresponds to the traditional setting where images are transmitted off-sensor and processed by a centralized processor. We will refer to this as the centralized approach, and use it as a comparison later.

\subsection{Distributed Formulation with a Gaussian Factor Graph}

\begin{figure}[t]
    \centering
    \includegraphics[width=0.66\linewidth]{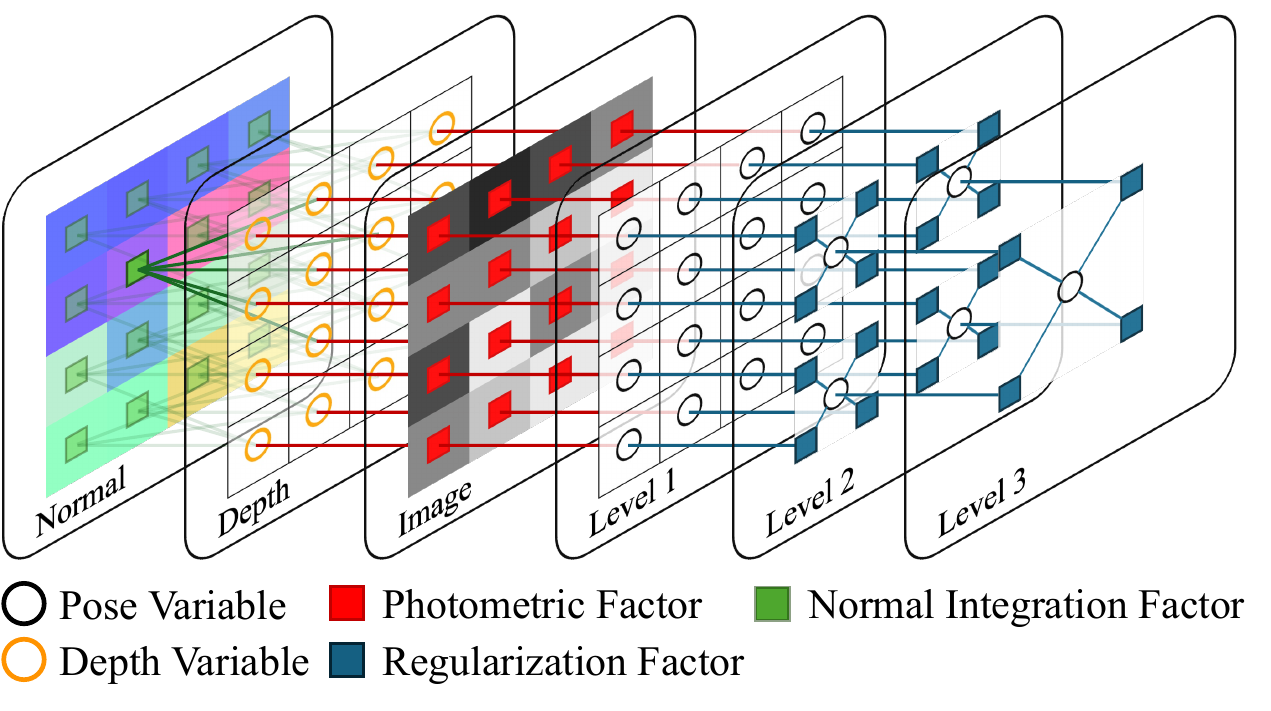}
    \caption{Topology of the proposed factor graph. The factor graph on the right side of the Image layer, related to camera pose estimation, follows the sharded topology of PixRO~\cite{alzugaray2024pixro}, with the domain extended to $\mathbb{SE}(3)$.
The factor graph on the left side of the Image layer reconstructs depth through the normal integration and photometric factors.
For the normal integration factors, note that all factors except the one located at the first row, first column are rendered semi-transparent to better illustrate the connection structure, whose topology is otherwise less comprehensible.}
    \label{fig:factorgraph}
    \vspace{-0.3cm}
\end{figure}

Our method is designed to implement the same direct visual tracking process but with computation and memory distributed across a pixel-parallel array of processing cores. We formulate the dense visual odometry and depth estimation problem as a
Gaussian factor graph~\cite{dellaert2017factor} describing the probabilistic relationship between variables of interest $v\in\mathcal{V}$ and factors $f\in\mathcal{F}$ which represent priors or measurements, such that the likelihood of the entire model is the product $p(\mathcal{V})=\prod_i{f_i(\mathcal{V}_{f_i})}$, where  $\mathcal{V}_{f_i}\subseteq\mathcal{V}$ is a subset of variables involved in each factor. 
This is the standard probabilistic formulation of a broad range of geometric estimation in SfM, SLAM, and robotics~\cite{dellaert2017factor}, and is the same formulation adopted by PixRO~\cite{alzugaray2024pixro} and BP-SF~\cite{scona2022scene}.

The key to our distributed implementation is that the stacked variable associated with each pixel can be stored locally at that pixel, and information about the factors that connect pixel variables can be stored altogether at the pixels at either end of the connection. As we will explain later, we implement optimisation to achieve marginal estimates of the variables via Gaussian Belief Propagation, which operates via local computation and message passing. Therefore no global memory access is needed by the pixels.

The overall structure of our factor graph for one timestep of visual odometry and depth estimation is shown in~\cref{fig:factorgraph}. A full spatiotemporal solution to sequential VO would involve a temporal stack of such graph fragments per timestep, connected by appropriate continuity factors. Here, we concentrate on explaining one timestep, where the goal is to estimate the camera motion from one image frame to the next, along with a depth map for the first frame.

At each pixel we store a stacked variable representing 3D camera motion and depth. 
We model 3D camera motion and rotation using $\mathbb{SE}(3)$ pose variables. Each pixel stores its own $\mathbb{SE}(3)$ estimate, but since we know that there is only one global motion of the whole camera, these variables are connected by strong Gaussian regularisation ``identity'' factors which encourage them to converge to the same value. We have experimented with different patterns that these factors could take, and we will discuss this later. \cref{fig:factorgraph} shows a hierarchical, sharded structure as was shown to have good performance in PixRO~\cite{alzugaray2024pixro}, while being able to implement even on existing PPAs with a few hops of direct-neighbor communication.

Each pixel has a photometric factor, which models the ``optical flow'' constraint that the camera motion and depth of a pixel determine with which pixel it should line up in the subsequent image frame, and that this corresponding pixel should have the same intensity.
Finally, we use factors that constrain the depth estimates of neighbouring pixels. These factors could simply model depth smoothness, but we have found it more effective to
use factors that constrain the depth of neighbouring pixels based on normal predictions coming from a local network, in the style of 
BINI~\cite{cao2022bilateral}, which showed that normal integration is highly suitable for pixel-level parallelization. We assume that accurate normal prediction from a small CNN is something that should be readily achievable in a pixel array in the near future, following work such as~\cite{Liu:etal:Frontiers2022}.
We will explain the details of this factor graph model shortly.

\subsection{Detailed Description of Factors}

In our distributed formulation, each pixel $p_i$ stores a stacked variable $\mu_i$ for each pixel $p_i$, whose first six elements $\mu_{i,[0:6]}$ are a local estimate of global camera motion, and whose final element $\mu_{i,[6]}$  is an up to scale estimate of the log-depth of the scene at that pixel. Our factor graph model has the following factors which constrain the absolute and relative values of these variables.
 
\noindent\textbf{Photometric factors:} This factor corresponds to the decomposition of the original centralized formulation in Eq.~\eqref{eq:photo} into local terms, each dependent on a single pixel, and represents the photometric difference between the corresponding pixel intensities of the two images:
\begin{equation}
E_{D}^i(\mu_i) = \frac{1}{2}\rho\left(\left\lVert\mathcal{I}_s\left[p_i\right] - \mathcal{I}_t\left[\mathcal{W}(p_i;\mu_i)\right]\right\rVert_{\Lambda_D}\right)^2\label{eq:distrb_photo}
\end{equation}
This energy is computed as each pixel reads the photometric value at its corresponding warped location in the target image, assuming its own camera pose estimate as the target camera pose.
On real PPA arrays, this can be implemented through local routing mechanisms that allow each processor to access neighboring pixel values.

\noindent\textbf{Prior factors:} This factor models a per-pixel prior hypothesis on the pose of the current camera.
On the first frame of a sequence, the prior pose is set to identity $[\mathbf{I};\mathbf{0}]$, serving as an assumption that the camera motion is slow, which stabilizes the optimization process—particularly during early iterations.

At later frames in a high frame-rate image sequence, this factor can propagate motion continuity priors from one timestep to the next.
\begin{equation}
    E_P^i(\mu_{i}) = \frac{1}{2}\left\lVert\mu_i\diamondminus\hat\mu_i\right\rVert_{\Lambda_P}^2\label{eq:prior}
\end{equation}

We also have factors that constrain the overall scale of the log depth at each pixel.
For the first frame in a sequence, we use a prior mean depth of 1, corresponding to a log-depth prior of 0. In subsequent frames, 
these factors can propagate depth estimates per pixel to future frames.

\noindent\textbf{Camera motion identity factors:} 
These are the crucial factors which encourage the multiple estimates of camera motion at each pixel towards a unique, global value. Without such factors, each pixel would find its own photoconsistent motion estimate, and in low-texture or repetitive scenes, these estimates would be highly inconsistent. 
\begin{equation}
    E_R^{i,j}(\mu_{i,[0:6]}, \mu_{j,[0:6]}) = \frac{1}{2}\left\lVert\mu_{i,[0:6]}\boxminus\mu_{j[0:6]}\right\rVert_{\Lambda_R}^2\label{eq:regul}
~.
\end{equation}

Since these factors are syncronizing a global variable, rather than having a local image interpretation, they can in principle be connected in any pattern. The simplest pattern would be to match the grid structure of pixels, with factors connecting all horizontally and vertically neighbouring pixel pairs. This is desirable from the point of view of matching the communication structure of current PPA hardware such as SCAMP, and we will continue to experiment further with this, but we currently find that it can require many iterations of GBP optimisation to achieve good global camera pose estimates with only these local connections. In the main experiments in this paper, we use a hierarchcal or sharded connection pattern for these identity factors.

\noindent\textbf{Normal integration factors:}
Finally, we use factors which help us to estimate smooth and consistent dense log depth across all pixels of the source image,
 the components $\mu_{i,[6]}$ of each pixel's variable $\mu_i$. Without such factors, which encode priors about the smoothness of the real world, accurate reconstruction would be possible only in extremely textured scene regions where photoconsistency alone might suffice.

Rather than assuming only that local scene depth changes smoothly, we use a stronger prior on the assumption that local estimates of surface orientation (normal map) are available. It is feasible to run small CNNs on a pixel-parallel array (e.g. \cite{Liu:etal:Frontiers2022, so2024pixelrnn}, and normal prediction can be performed extremely well with CNNs~\cite{bae2024rethinking} as the scene normals are relatively local properties, rather than global quantities such as scene depth.
Previous work such as 
BINI~\cite{cao2022bilateral} has shown that local integration of a normal map from a CNN can produce accurate local depth estimates up to scale within regions without depth discontinuities.

Our factors constrain the depth difference between neighbouring pixels to be consistent with the locally measured normals:
\begin{equation}
\begin{aligned}
    E_{N_x}^{i>j}(\mu_{i,[6]}, \mu_{j,[6]}) &= \frac{1}{2}\left\lVert (\mu_{i,[6]}-\mu_{j,[6]})+\frac{n_x}{\tilde{n}_z}\right\rVert_{\Lambda_N}^2\\
    E_{N_y}^{i>j}(\mu_{i,[6]}, \mu_{j,[6]}) &= \frac{1}{2}\left\lVert (\mu_{i,[6]}-\mu_{j,[6]})+\frac{n_y}{\tilde{n}_z}\right\rVert_{\Lambda_N}^2,
    \label{eq:gbpni}
\end{aligned}
\end{equation}where the upper equation applies to horizontally adjacent pixels and the lower one to vertically adjacent pixels.
$\tilde{n}_z$ denotes the $z$-component of the normal scaled to the image-plane coordinate system (see p.~4 of BINI~\cite{cao2022bilateral} for the exact definition), while $n_x$ and $n_y$ denote the $x$ and $y$ components of the surface normal, respectively.

\subsection{Iterative Gaussian Belief Propagation}

To achieve camera motion and depth estimation, we need to rapidly and incrementally optimise this factor graph model to obtain marginal estimates of the camera motion and depth variables. The desire to implement this on a pixel-parallel array rules out standard algorithms and software libraries which assume that all variables and factors are jointly accessible. Instead, we use Gaussian Belief Propagation (GBP), which can obtain marginal estimates via fully distributed computation by passing Gaussian messages locally between factors and variables. Assuming Gaussian distributions, the messages admit analytic update rules, allowing nodes and factors to update locally without explicit synchronization. In the following, we summarize the steps tailored to the pose and depth of the camera; fuller details appear in~\cite{alzugaray2024pixro, ortiz2021visual, davison2019futuremapping, murai2024distributed, murai2023robot}.

In our method, each pixel stores a stacked variable $v \in \mathcal{V}$  which represents camera pose and log-depth in $\left<\mathbb{SE}(3), \mathbb{R}\right>$. These variables are generally related through nonlinear factors; thus, the linearization point must be progressively approximated. Following the conventions in~\cite{alzugaray2024pixro, murai2023robot, murai2024distributed}, we take the linearization point of each pose-log-depth variable to be its current mean, denoted by $\bar{\mu} \in \left<\mathbb{SE}(3), \mathbb{R}\right>$. The corresponding probabilistic variable is then defined as:
\begin{equation}
v = \bar{\mu} \diamondplus \presup{\bar{\mu}}{\xi}, \quad \text{where} \quad \presup{\bar{\mu}}{\xi} \sim \mathcal{N}(0, \Sigma_v)
\label{eq:var}
~.
\end{equation}

Here, $\presup{\bar{\mu}}{\xi} \in \langle \mathfrak{se}(3), \mathbb{R} \rangle$ denotes a stacked variable consisting of a Lie-algebra element and a scalar component. We regard this as an element of a Euclidean space, exploiting the vector-space structure of the Lie algebra to apply standard probabilistic methodologies. Throughout this work, we adopt the Lie-group notation of~\cite{sola2018micro} for the direct product $\langle \mathbb{SE}(3), \mathbb{R} \rangle$: in particular, $\oplus$ denotes a pose update by composing with the exponential map, and $\diamondplus$ denotes the component-wise extension of $\oplus$ to the composite manifold.

The residual of any factor involving such variables can then be approximated by a first-order Taylor expansion similar to~\cite{alzugaray2024pixro} as follows:
\begin{equation}
\begin{aligned}
r(\mu) &= r(\bar{\mu} \diamondplus \tau) \approx r(\bar{\mu}) + \frac{Dr}{D\mu} \bigg|_{\mu = \bar{\mu}} \tau\\
&= r(\bar{\mu}) + J(\bar{\mu})\tau = \bar{r} + \bar{J}\tau
~,
\end{aligned}
\end{equation}
where $J(\cdot)$ denotes the right Jacobian of $\left<\mathbb{SE}(3), \mathbb{R}\right>$. Note that the Jacobian matrix is block diagonal, as this composite is a direct product group.

Then, each energy term can be accordingly approximated as follows:
\begin{equation}
\begin{aligned}
E(\tau) &= \frac{1}{2}||r(\bar{\mu}\diamondplus\tau)||^2_{\Lambda_r}\approx||\bar{r}+\bar{J}\tau||^{2}_{\Lambda_r}\\
&\propto \frac{1}{2}\tau^T\bar{J}^T\Lambda_r\bar{J}\tau + (\bar{J}^T\Lambda_r\bar{r})^T\tau
~.
\end{aligned}
\end{equation}
We follow the standard steps of GBP to send factor-to-variable messages, variable-to-factor messages, to relinearize factors and to compute variable beliefs, as in \cite{alzugaray2024pixro,murai2023robot,murai2024distributed}.

\subsection{Keyframe-based Tracking Strategy}
Direct photometric tracking method is well known to have a limited basin of convergence around the optimum, and our method is likely to converge to sub-optimal camera pose and depth when the inter-frame motion is large. 
In a PPA, on-board processing means there is no need to transmit video data and the frame-rate can be kept high so motions from one frame to the next can be kept small.
However, we have found that simply running very high-rate frame-to-frame 6DoF odometry and naively integrating these estimates over time is also problematic: if the motion between the reference and target frames becomes excessively small, it becomes difficult to reliably disambiguate rotation and translation, degrading tracking accuracy. We therefore introduce a local keyframing strategy. Specifically, we treat reference frame as a fixed keyframe during tracking for a number of steps, while the target frame is updated incrementally as camera motion accumulates, and 6DoF poses are always estimated locally relative to this keyframe.
In the following sections, we analyse the characteristic advantages and limitations of this keyframe-based formulation.

\section{Experimental Evaluation}

\begin{figure}[t]
    \centering
    \begin{subfigure}{0.8\linewidth}
        \centering
        \includegraphics[width=\linewidth]{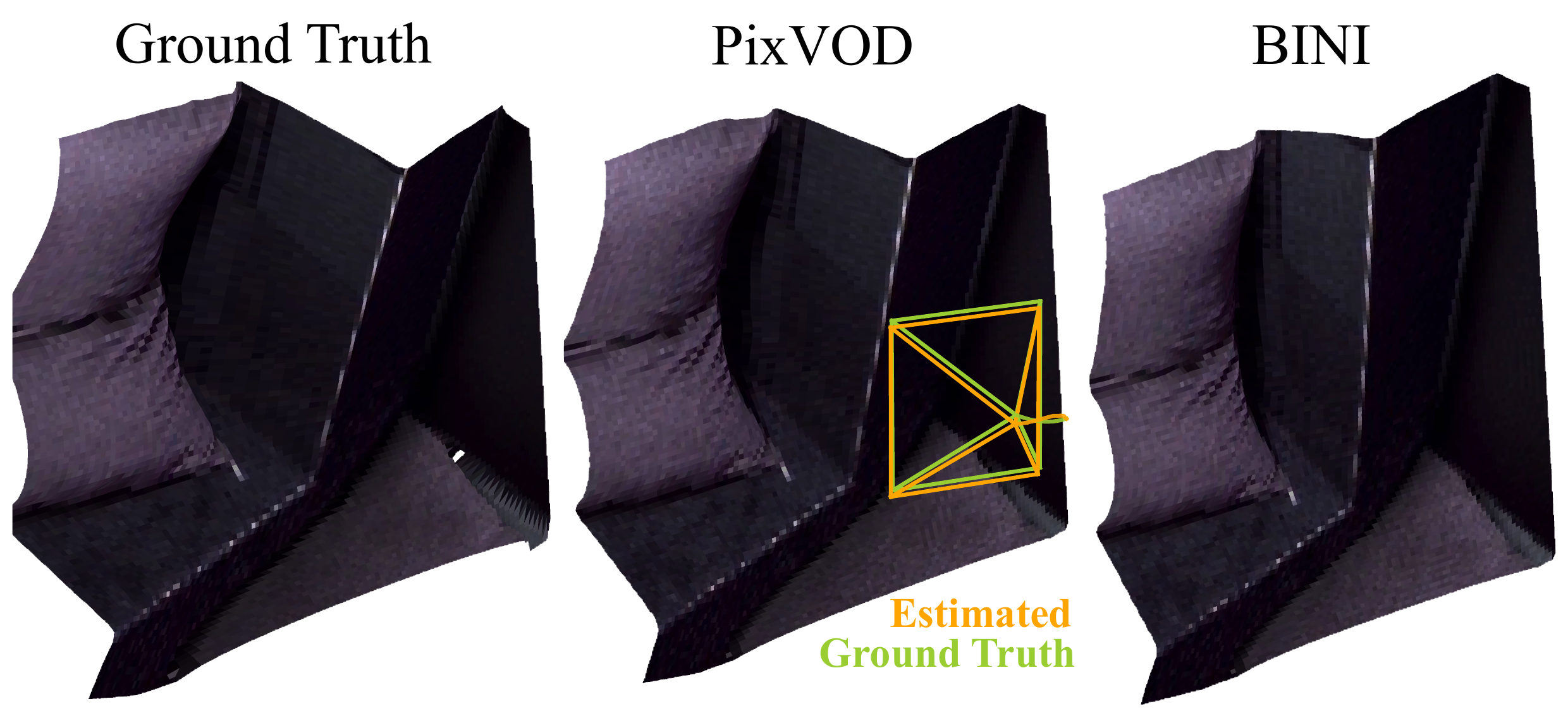}
        \label{fig:depth_traj_a}
    \end{subfigure}


    \begin{subfigure}{0.8\linewidth}
        \centering
        \includegraphics[width=\linewidth]{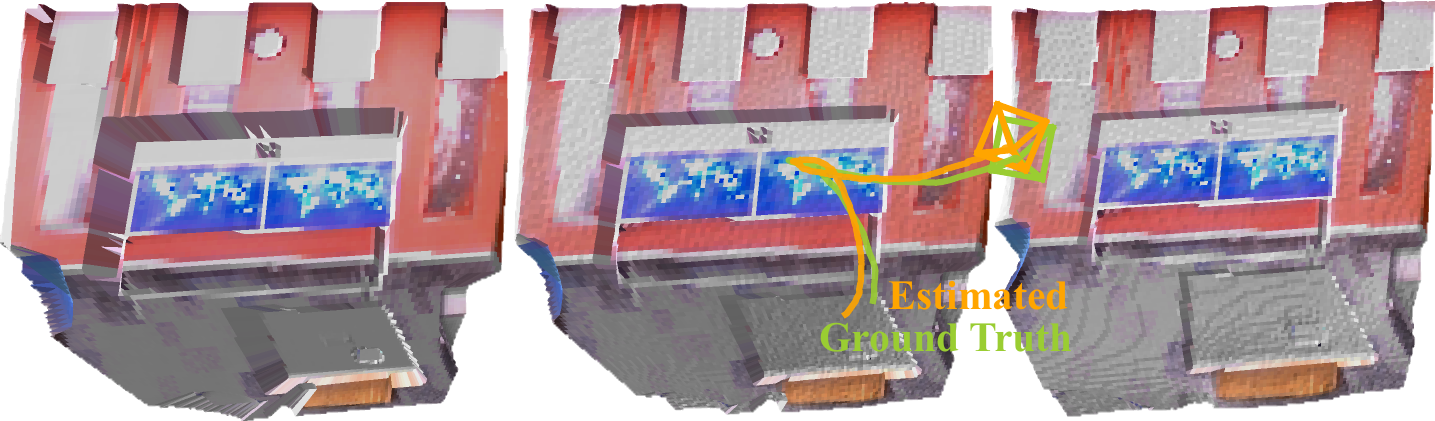}
        \label{fig:depth_traj_b}
    \end{subfigure}

    \caption{Visualization of an estimated local trajectory and depth with \textbf{(top)} relatively small motion and being close to surface, \textbf{(bottom)} relatively large motion and being far from surface. The view frustum of the camera is visualized only on the final pose. For the top image, the trajectory from a reference keyframe to the next one is shown, and for the bottom trajectory, 5 keyframes are in between, while omitted for visibility of the trajectory. Unprojected depth was connected to form a mesh and textured with an image for easier comprehension. Note that the scale of the estimated depth of the proposed method is more accurately reconstructed compared to when BINI is solely used, as explained in \cref{sec:photoni}.}
    \label{fig:depth_traj}
    \vspace{-0.4cm}
\end{figure}


\subsection{Implementation Details}

As in most publicly available GBP implementations~\cite{davison2019futuremapping, ortiz2021visual}, we alternate synchronously between the message passing stage and the belief update stage, iterating over both factor and variable nodes.
Only the variable and factor-to-variable messages are stored explicitly.
At each relinearization step, these variables and messages are projected onto the tangent space of the manifold at the updated linearization point.
We employ the Huber function as a robust loss, which is a common choice in photometric optimization, and use the covariance weighting method from~\cite{davison2019futuremapping}
in every GBP factor message passing step.



Our current implementation is a GPU simulation of a potential future implementation on a true pixel parallel array. 
All steps of the GBP algorithm are parallelized using JAX, and the experiments were conducted on a desktop machine equipped with an NVIDIA RTX 4090 GPU and an AMD Ryzen 9950X CPU.
\subsection{Evaluation Details}\label{sec:eval_details}

For the proposed experiments, we generate sequences of images of 128 by 128 pixels from the publicly available Replica dataset~\cite{straub2019replica}, which provides photorealistic 3D reconstructions of real-world environments.
We use the sequences \textit{office\_\{0, 1, 2, 3\}} adopted from iMap~\cite{sucar2021imap} and Nice-SLAM~\cite{zhu2022nice}, but upsample each frame-to-frame interval by a factor of 10 to simulate the small baselines that would arise at high frame-rate in a PPA.
For trajectory upsampling, we employ ScLERP~\cite{kavan2006dual}, an extension of the linear interpolation method SLERP~\cite{shoemake1985animating} from $\mathbb{SO}(3)$ to $\mathbb{SE}(3)$.
Unless otherwise specified, ground-truth surface normals are used. All metrics on translation and depth are evaluated after the scale of estimation is aligned with that of ground-truth, by mimizing $L_2$ error.


Our hyperparameters are $\sigma_D, \sigma_P, \sigma_R, \sigma_N,$ and $t_{\mathrm{huber}}$, which are respectively used in the form $\Lambda = \frac{1}{\sigma^2} I$ and as a Huber threshold, and can be interpreted as controlling the assumed noise level of the underlying model (or simply, strength) of each factor. We empirically determine suitable magnitudes for these hyperparameters, and, unless otherwise specified, we set $(\sigma_D, \sigma_P, \sigma_R, \sigma_N, t_{\mathrm{huber}}) = (5\times10^{-3}, 1, 4\times10^{-4}, 10^{-3}, 400)$ at the leaf level of our factor graph. The values of $\sigma_P$ and $\sigma_R$ are halved at each higher level of the sharded quadtree factor graph in \cref{fig:factorgraph} compared to the level below. This choice reflects the observation that lower levels have access only to more local information and therefore produce pose estimates that are less reliable than the globally aggregated poses at higher levels. We empirically validate this design choice in the supplementary material.



\subsection{Analysis on Keyframe-based Tracking}~\label{sec:keyframe}
\begin{figure*}[t]
    \centering
    \includegraphics[width=1\linewidth]{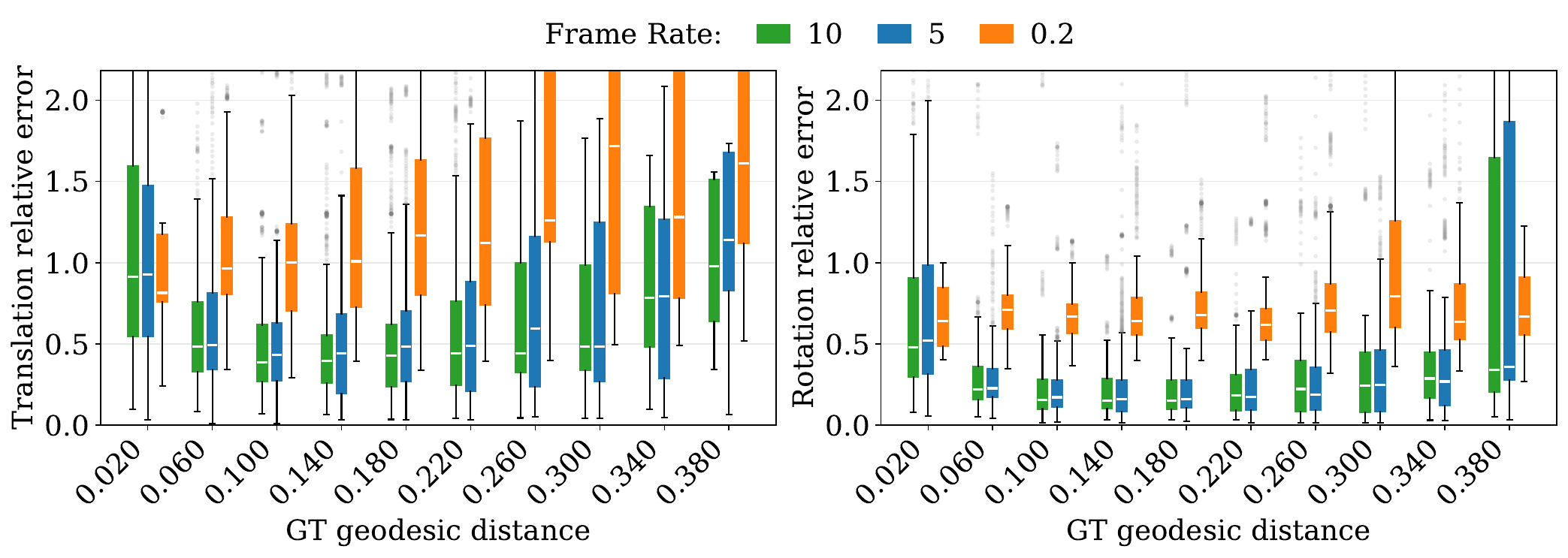}
    \caption{Comparison of estimated pose error across various frame rates, based on the deviations of the target camera pose from origin $[\mathbf{I};\mathbf{0}]$. As the pose change between consecutive frames increases, i.e., as the frame rate decreases, the convergence basin of the proposed method (with respect to the deviation of the target camera pose from the origin) becomes smaller, and the corresponding error increases. However, the benefit of reduced frame-to-frame motion exhibits diminishing returns (cf. framerates {\color{tabgreen}\textbf{10}} and {\color{tabblue}\textbf{5}}), similar to which has also been observed in related domains such as event-based vision~\cite{handa2012real}.} 
    \vspace{-0.4cm}
    \label{fig:error_wrt_framerates}
\end{figure*}
In this section, we analyze the choice of keeping a reference frame and optimizing with respect to a continually updated target frame. Along the upsampled Replica trajectory, we sampled 10 keyframes per scene. In each keyframe, we extract ground-truth surface normals and use them as the prior for normal integration (Eq.~\eqref{eq:gbpni}); for the subsequent 300 frames, we treat each as a target frame from which observations for the photometric factor (Eq.~\eqref{eq:distrb_photo}) are sampled. For every target frame, we perform 100 synchronized GBP iterations, where one synchronized iteration applies message passing and belief update sequentially to all factors and variables. The choice of 100 iterations is the minimal count that suffices to reach a convergent solution for almost all target frames. At the end of the optimization loop for each target frame, the pose estimations of all pixels are averaged to form the final estimation for that frame. When moving on to the next frame, all variables are initialized with the solution from the previous frame.

Here, we analyze why keyframe–frame optimization is essential by varying the update frequency of target frames to enlarge inter-frame visual change, and comparing against the original update rate. (\cref{fig:error_wrt_framerates}) Specifically, we follow the protocol mentioned above (framerate 10); we update the target frame every two frames while doubling the number of optimization iterations per target frame (framerate 5); and we update the target frame every fifty frames while performing fifty times as many iterations per target frame (framerate 0.2). Using the converged solution at the end of each target-frame optimization as a data point, we obtain the box-and-whisker plot in \cref{fig:error_wrt_framerates}. The x-axis reports the geodesic distance of ground-truth, defined as the $l_2$-norm of the $\mathfrak{se}(3)$ vector of the relative pose of ground-truth with respect to the pose of the keyframe; the y-axis shows the relative error of our estimate: $||t_{\mathrm{est}}-t_{\mathrm{gt}}||_2/||t_{\mathrm{gt}}||_2$ for translation (where $t$ is taken from translation part of $\mathbb{SE}(3)$), and the axis-angle vector for rotation.

As shown in \cref{fig:error_wrt_framerates}, for framerate 0.2 the method settles more frequently at suboptimal solutions as the target-frame updates become infrequent. However, comparing framerate 10 to 5 indicates that the benefit of further increasing the update frequency is limited; in real-world settings, this limit would be accentuated by the reduced signal-to-noise ratio caused by shorter exposure times. The plot also offers insight into when camera-pose tracking is particularly challenging: when motion is too small, it becomes difficult to disambiguate the motion type (e.g. vertical translation versus rotation about a horizontal axis), whereas when motion is too large, the region providing usable visual cues diminishes, making accurate pose tracking more difficult.




\subsection{Comparison of Pixel-Parallel and Centralized Visual Odometry Estimation}
\begin{figure*}[t]
    \centering
    \includegraphics[width=1\linewidth]{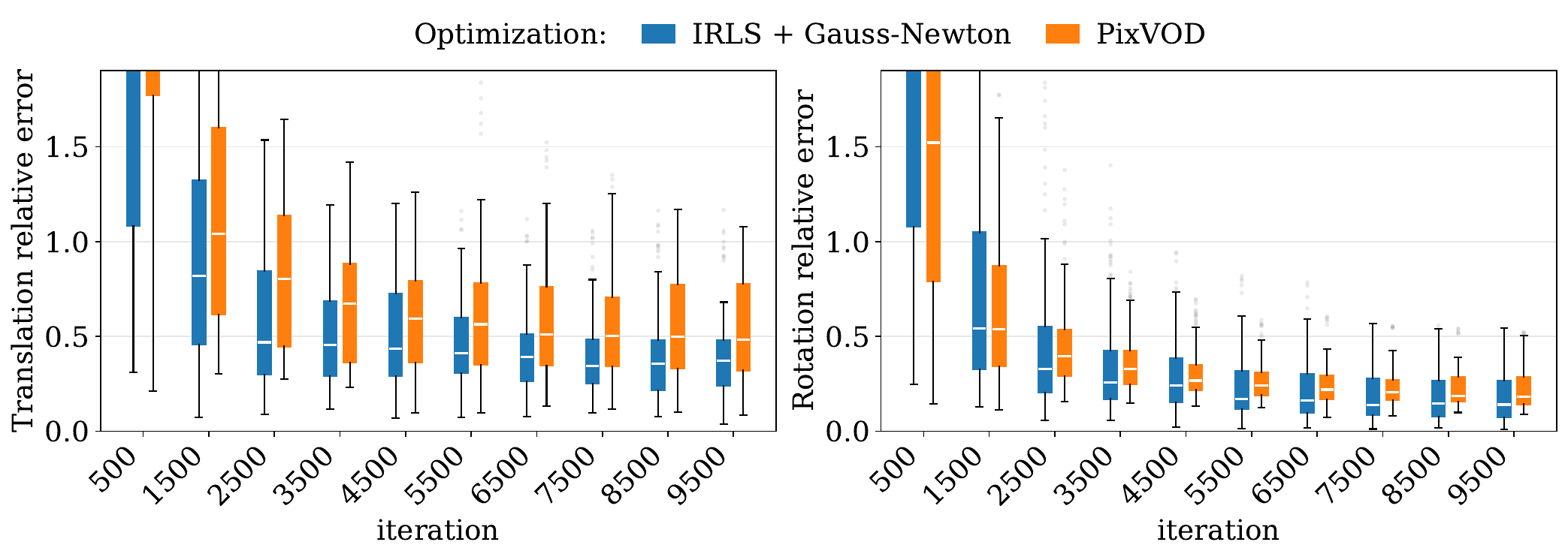}
    \caption{Comparison of the proposed method against the baseline method, which is solved in a centralized manner. We formulated the centralized optimization problem as an Iterative Reweighted Least Squares (IRLS), and solved it with the Gauss-Newton method.} 
    \label{fig:centralized_vs_pixvod}
    \vspace{-0.3cm}
\end{figure*}

We compare the optimization performance of the proposed GBP-based method against the aforementioned centralized baseline. The centralized approach (\cref{sec:centralized}) optimizes a global estimator of pose and depth with simultaneous access to all pixel measurements, and is solved using Gauss–Newton. Since the problem is nonlinear as proposed in Eq.~\eqref{eq:photo} and~\cite{cao2022bilateral}, we model it via iteratively reweighted least squares (IRLS), equally weighting two objectives. Following the earlier experiments, we set the first frame as the reference and treat the subsequent 100 frames, over which convergence occurs in the vast majority of cases statistically, as continuously updated target frames; we perform 100 iterations of optimization per frame in sequence. 

As shown in Fig.~\ref{fig:centralized_vs_pixvod}, the centralized method converges faster and to higher accuracy than our approach, as expected; however, this presumes per-iteration access to global pixel information. In contrast, the proposed method accesses pixels only locally and independently at each variable, offering an alternative optimization strategy in environments where global access is atypically expensive.




\subsection{Photometrically Guided Normal Integration}\label{sec:photoni}

\begin{figure}[t]
    \centering
    
    \begin{subfigure}[t]{0.475\linewidth}
        \centering
        \includegraphics[width=\linewidth]{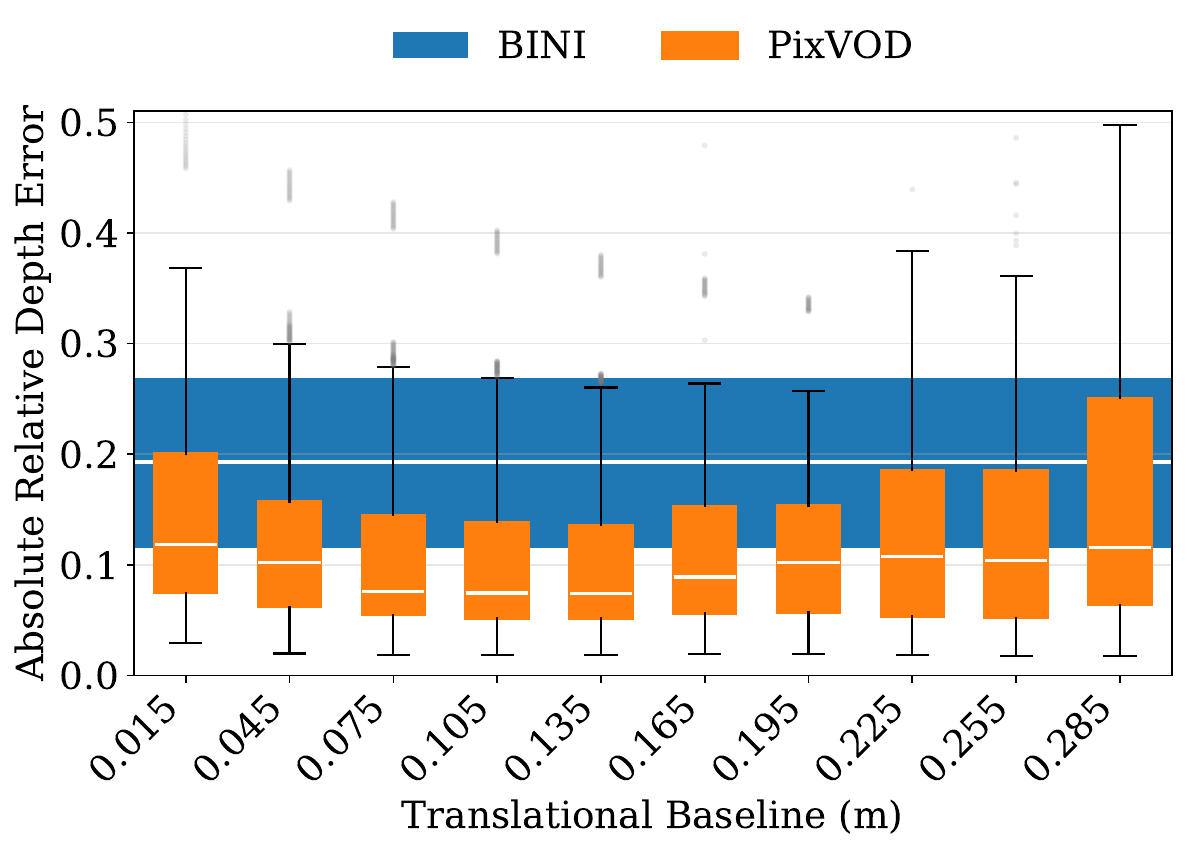}
        \caption{}
        \label{fig:abs_rel_depth}
    \end{subfigure}
    \hfill
    \begin{subfigure}[t]{0.5\linewidth}
        \centering
        \includegraphics[width=\linewidth]{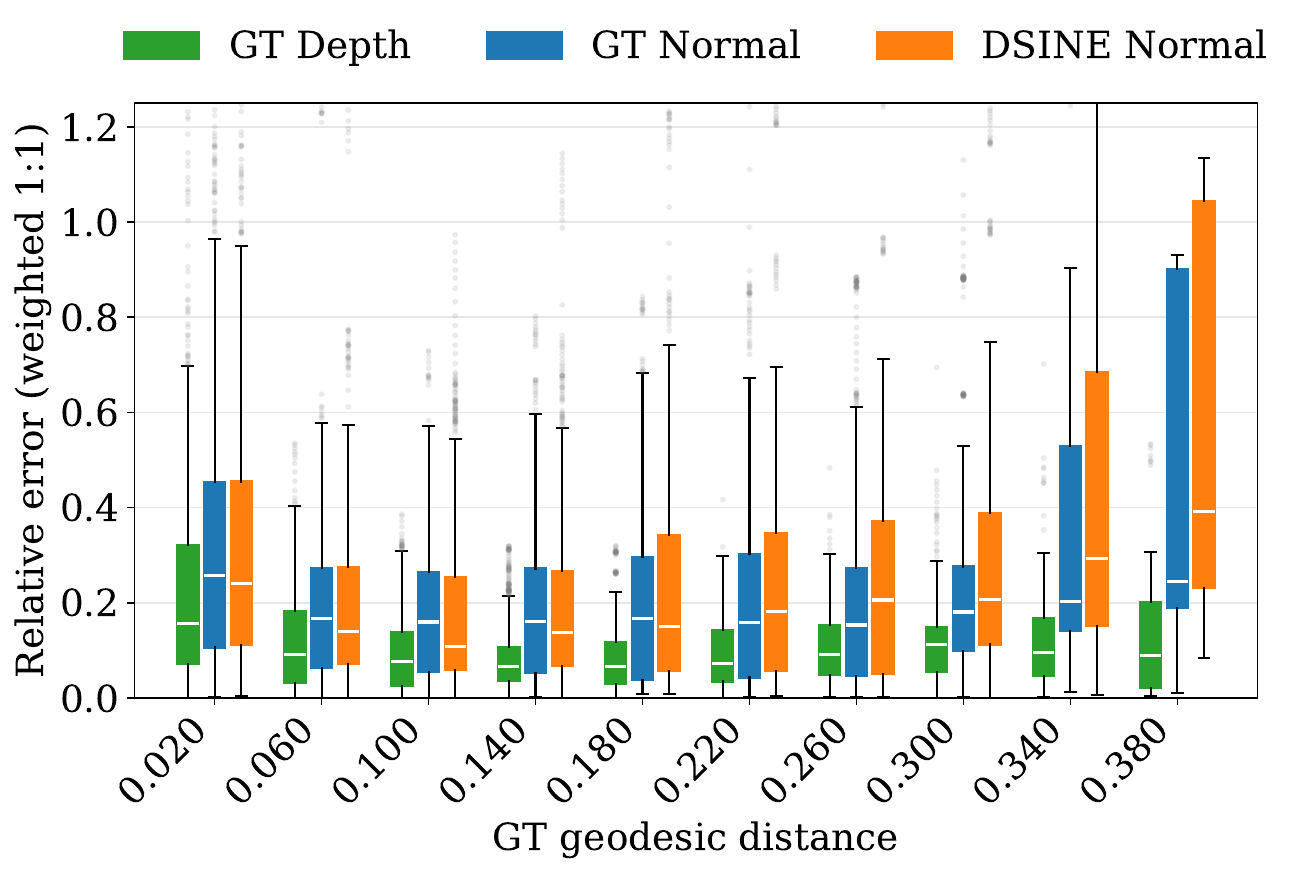}
        \caption{}
        \label{fig:geometric_prior_comparison}
    \end{subfigure}
    
    \caption{\textbf{(a)} Comparison of reconstructed depth of BINI~\cite{cao2022bilateral} after convergence against our reconstructed depth with various translational baselines of camera. \textbf{(b)} Evaluation on camera poses estimated with various structural priors. Note that most of the experiments reported in this paper are paired with the Ground Truth Normals.}
    \vspace{-0.4cm}
\end{figure}
In this section, we compare our method against pure normal integration without photometric guidance. In line with recent work~\cite{zhu2024nicer,mazur2024superprimitive}, the results indicate that incorporating photometric correspondences can extend normal-integration-based reconstruction beyond the object level, by guiding both sides of depth-discontinuous boundary to be pulled apart.

As shown in \cref{fig:abs_rel_depth} and exemplified in \cref{fig:depth_traj}, when initialized with the depth at BINI’s convergence, our approach achieves substantially higher depth quality than BINI after its own optimization. Note that whiskers are omitted for the BINI results. Consistent with \cref{sec:keyframe}, there exists a range of camera translations for which the depth reconstruction gains are most pronounced.

\subsection{Ablation Study on Geometric Prior}

In \cref{fig:geometric_prior_comparison}, we evaluate the proposed method with a variety of structural priors to verify that its pose estimation remains robust irrespective of the prior type. As expected, using ground-truth depth yields the best convergence. Interestingly, when using DSINE~\cite{bae2024rethinking} normals, the quality of the converged pose is not substantially worse than with ground-truth normals, aside from a reduced basin of convergence. This indicates that our approach generalizes to real-world use when paired with practical normal-estimation methods.

\section{Conclusions}

We present the first per-pixel distributable algorithm for jointly optimizing camera pose and depth. Our approach demonstrates that each pixel-processor can infer global image properties using only its own pixel measurement, a local geometric prior, and message passing with neighboring processors. We also provide an effective strategy for camera tracking in the typical operating regime of PPAs, high frame rates with small inter-frame changes, and analyze the operating range and limitations of the method, providing insights that can inform the design of next-generation PPAs. 

In the future, we will investigate whether gradual weakening of the inter-pixel factors that enforce a single global camera motion estimate, and so to what extent this will enable accurate non-rigid 3D scene flow estimation. We also plan to continue to experiment with different patterns of inter-pixel factor connections, to find the best compromise between patterns that enable rapid optimisation and those which are feasible and efficient to design into real future chips.

\section*{Acknowledgements}
This research is supported by the EPSRC grant EP/Y020499/1. We appreciate all members of the Robot Vision Group for insightful discussions.

%
%
\bibliographystyle{splncs04}
\bibliography{arxiv}

\clearpage

\appendix

\section{Qualitative Results}
In the second part of the accompanying video \textit{pixvod.mp4}, we demonstrate how the proposed method accumulates a trajectory in the Replica scene. Starting from the first frame as the keyframe, we process 150 frames with the procedure discussed in Sec.~4.3, designate the 151st frame as a new keyframe, and then repeat this cycle---each time extending the trajectory. We proceeded in a total of 2000 frames.

For visualization, the depth estimated at each keyframe is unprojected in the coordinate frame of its corresponding keyframe, meshed, and then transformed into the global frame (defined by the first keyframe). Since the depth optimized at the final frame of each keyframe segment is always retained in the global frame, the unprojected-depth visualizations accumulate over time, which can be observed clearly in the video.

\section{Analysis on Computational and Communicational Cost}
\subsection{Analysis on Inter-pixel Communication Pattern}\label{sec:compattern}
We provide a comparison on total computation, communication, and memory costs of hierarchical and grid structures of camera motion identity factors, assuming multiple executions of neighborwise message passing to handle multi-hop interactions. 
The right-hand table of~\cref{tab:VO} shows FLOPs per pixel for factor calculations, variable-to-factor and factor-to-variable messages, the latency of message passing (assuming 1 floating-point bandwidth), the total, and the number of parameters per-pixel. Although the quad-tree takes $\times1.5$ more FLOPs, it is still far more efficient, since the grid, due to its loopiness, needs at least 50 times more GBP iterations. That the loopiness of flat grid topology causes the late convergence of GBP and thus is computationally more expensive aligns with the insight from PixRO~\cite{alzugaray2024pixro}.

\subsection{Approximated Runtime Analysis on SCAMP-5}\label{sec:approxcomp}
To demonstrate that the computational cost of the proposed method is not far from what contemporary PPA can give, the conversion from computational cost in FLOPs to the number of cycles of contemporary PPAs is provided here.

In a bit-serial manner of floating point calculations, binary operations can achieve 96 cycles/FLOP for FP16 representation~\cite{ahlander1992floating}. This is under the assumption that each binary operation can be done in a cycle. This enables SCAMP-5, a well-known PPA that conducts binary operation per cycle per pixel-processor while running at 10MHz, to achieve 104.2KFLOPs/px/s. Taking the total number of FLOPs/px needed for each iteration of GBP from the right-hand table of \cref{tab:VO}, and considering that 100 iterations are needed for GBP to converge, SCAMP-5 can yield 1-2FPS for the proposed method. The SCAMP series is optimized for analog operations, leaving immediate room for speedups with floating-point support.


\begin{table}[t]
\centering
\small
\setlength{\tabcolsep}{5pt}

\caption{
    \textbf{(Left)} Comparison on performance and amount of computation against SoTA depth-guided visual odometry solver, MADPose\cite{yu2025relative}. \textbf{(Right)} Comparison on varied topologies of factor subgraph of camera motion identity factors.
}

\begin{minipage}{0.59\columnwidth}
\centering
\begin{tabularx}{\linewidth}{@{}p{0.2em}X|ccc@{}}
\toprule
& Prior & DSINE & \multicolumn{2}{c}{DepthAnythingV2} \\
& Match & GBP & ORB+NN & SP+LG \\
\midrule
\multirow{3}{*}{\rotatebox[origin=c]{90}{\scriptsize AUC(\%)}} 
& @\phantom{1}$5^{\circ}$ & \phantom{1}3.29 & \phantom{1}4.71 & 21.87 \\
& @$10^{\circ}$           & 13.84 & 13.27 & 41.28 \\
& @$20^{\circ}$           & 34.25 & 29.60 & 61.11 \\
\midrule
\multirow{2}{*}{\rotatebox[origin=c]{90}{\scriptsize FLOP}}
& Prior & 7.27G & 35.0G & 35.0G \\
& Match & 558M & 230M & 6.06G \\
\bottomrule
\end{tabularx}
\end{minipage}
\hfill
\begin{minipage}{0.40\columnwidth}
\centering
\begin{tabularx}{\linewidth}{@{}p{0.2em}X|cc@{}}
\toprule
& & Hierarchical & Grid \\
\midrule
\multirow{5}{*}{\rotatebox[origin=c]{90}{\scriptsize FLOPs/px}}
& Factor & 225 & 199 \\
& V2F & 224 & 168 \\
& F2V & 968 & 726 \\
& Latency & 294 & 42 \\
& Total & 1711 & 1135 \\
\midrule
& Lineari. & \multicolumn{2}{c}{1525 (FLOPs/px)}\\
\midrule
\multirow{1}{*}{\rotatebox[origin=c]{90}{\scriptsize /px}}
& Params & 75 & 56 \\
\bottomrule
\end{tabularx}
\end{minipage}

\label{tab:VO}
\vspace{-0.2cm}
\end{table}

\subsection{GPU Simulation of PixVOD Compared to VO Methods}

Wallclock time and FLOPs compared with other methods are given below. However, our workload is poorly-suited to GPU: $\mathbb{SE}(3)$ linearization of messages and nodes lead to fragmented access to global memory, contributing 94\% of wall-clock time but only 28\% of the FLOPs. Also, JAX is not tailored to factor graphs and GBP optimization, and is thus inefficient for our implementation. We estimate DTAM~\cite{newcombe2011dtam}'s FLOPs from the paper. Time and FLOPs are averaged across frame-pairs on our Replica sequences. As DROID-SLAM~\cite{teed2021droid} (DSLAM) and DTAM are well-optimized for GPU, they run faster despite more FLOPs, which further motivates on-sensor processing that exploits computation locality.

\begin{table}[h]
\centering
\setlength{\tabcolsep}{4pt}

\caption{
Wallclock time and FLOPs compared with DSO~\cite{engel2017direct}, DROID-SLAM (DSLAM)~\cite{teed2021droid}, and DTAM~\cite{newcombe2011dtam}.
}

\resizebox{0.75\linewidth}{!}{\begin{tabular}{l|c c c c }  
\toprule
     & Ours+DSINE & DSO & DSLAM & DTAM (est.)\\
    \midrule
    Hardware & GPU (\textcolor{red}{Simulation}) & CPU & CPU+GPU & CPU+GPU \\
    Walltime (ms) & 853 & 13.2 & 16.5 & -- \\
    FLOPs & 631M & 33.9M & 25.6G & 870M \\ 
\bottomrule
\end{tabular}}
\label{tab:runtime}
\end{table}

To distinguish the proposed method against DTAM in both parallelization structure and efficiency, DTAM's per-pixel parallelization plus reduction is comparably parallel for pose estimation. The mapping part of DTAM, however, must maintain an explicit $M\times N\times S$ cost volume: a per-frame build plus a depth-wise $\alpha$-search across all $S$ hypotheses. Our pipeline replaces this with GBP-based normal integration on keyframes, with a keyframe-to-frame ratio of $\approx$300, resulting in the amortized normal estimation plus integration factor to take only $<$6.5\% of our per-frame budget (\cref{sec:practicalnormal}).

\section{Discussion}
\subsection{Practicality of Assuming Normal Estimates}\label{sec:practicalnormal}
Although shallow CNNs and RNNs have recently been shown to run on-sensor on modern PPAs~\cite{so2024pixelrnn}, the large CNN-based model used to obtain our geometric prior~\cite{bae2024rethinking} may exceed the capacity of current PPA hardware. Nevertheless, our method is intentionally structured to amortize this cost over time by applying the prior network only to sparse keyframes, which remain fixed over a few hundred target frames. This makes it possible to execute DSINE~\cite{bae2024rethinking} in a time-interleaved fashion, loading only a few layers at a time while GBP iterations continue on the incoming target frames. 

Quantitatively, if the DSINE computation is amortized across 10 target frames, its cost amounts to approximately $11$ KFLOPs per pixel per frame, according to \cref{sec:approxcomp} and \cref{tab:VO}: $\frac{7.27\text{ GFLOPs}}{10\text{ frames}\times(256\times256)\text{ px}} \approx 11\text{ KFLOPs}$. This is substantially less (approximately 6.5$\%$) than the cost of the proposed GBP inference, which is approximately $171$ KFLOPs per pixel per frame, which is given by $1711$ FLOPs per pixel per iteration $\times 100$ iterations.

A further potential concern is the feasibility of alternating between different types of computation, such as GBP iterations and neural network forward passes. However, such temporal interleaving is a fundamental operating principle of modern PPAs, as demonstrated in~\cite{bose2025descriptor}.


\subsection{Evaluation Details}
This section describes the relative error metric used throughout our evaluations. In the following equations, we present, from top to bottom, the computation methods for the Translation Relative Error, the Rotation Relative Error, and the overall Relative Error (with a 1:1 weighting), corresponding to those used in Sec.~4.4–4.6 of the main paper. For the Lie-algebra operations, note that we are following the notation of~\cite{sola2018micro}.
\begin{equation}
\begin{aligned}
    e_{\text{translation}} =& \frac{||t_{\rm{est}}-t_{\rm{gt}}||_2}{||t_{\rm{gt}}||_2},\\
    e_{\text{rotation}} =& \frac{\left\lVert\rm{Log}\left(\rm{Exp}(\theta_{\rm{est}})\rm{Exp}(\theta_{\rm{gt}})^{-1}\right)\right\rVert_2}{||\theta_{\rm{gt}}||_2},\\
    e_{\text{pose}}
    \label{eq:rel_err} =& \frac{\sqrt{\left\lVert\rm{Log}\left(\rm{Exp}(\theta_{\rm{est}})\rm{Exp}(\theta_{\rm{gt}})^{-1}\right)\right\rVert_2^2 + \left\lVert t_{\rm{est}}-t_{\rm{gt}}\right\rVert_2^2}}{\sqrt{||\theta_{\rm{gt}}||_2^2+||t_{\rm{gt}}||_2^2}}
\end{aligned}
\end{equation} where,
\begin{equation}
    \rm{Exp}\left(\begin{bmatrix}
    \theta\\\rho
    \end{bmatrix}\right) = \begin{bmatrix}
    \rm{Exp}(\theta) & \mathbf{RV}(\theta)\rho \\
    \mathbf{0} & 1
    \end{bmatrix} = \begin{bmatrix}
    \rm{Exp}(\theta) & t \\
    \mathbf{0} & 1
    \end{bmatrix}.
    \label{eq:exp_err}
\end{equation}

Note that, $d_{\rm{gt}}=\sqrt{||\theta_{\rm{gt}}||_2^2+||t_{\rm{gt}}||_2^2}$ is what we call ``GT geodesic distance'' in the main paper and graphs. The estimated pose as $\mathfrak{se}(3)$ element $\xi_{\rm{est}}=\left[ \begin{smallmatrix}\theta_{\rm{est}}\\\rho_{\rm{est}}\end{smallmatrix} \right]$ is calculated as an average of pose estimations at the lowest level of the factor graph topology:
\begin{equation}
    \xi_{\rm{est}}=\frac{1}{N}\sum_{i=1}^N{\mu_{i,[0:6]}},
    \label{eq:estimator_avg}
\end{equation} where $N$ is the number of pixels.


\begin{figure}[t]
\centering
\begin{minipage}[c]{0.42\columnwidth}
\centering
\vspace{0.5cm}
\includegraphics[width=\linewidth]{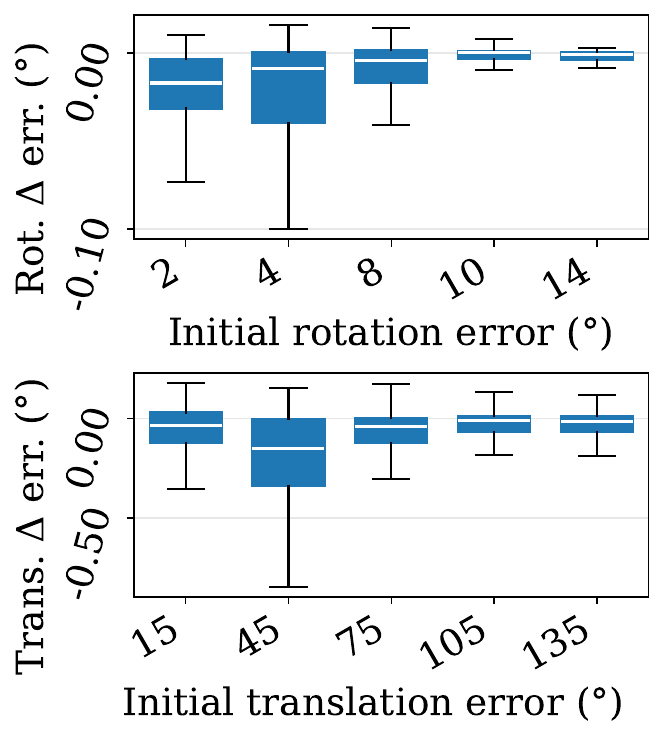}
\end{minipage}
\begin{minipage}[c]{0.41\columnwidth}
\centering
\includegraphics[width=\linewidth]{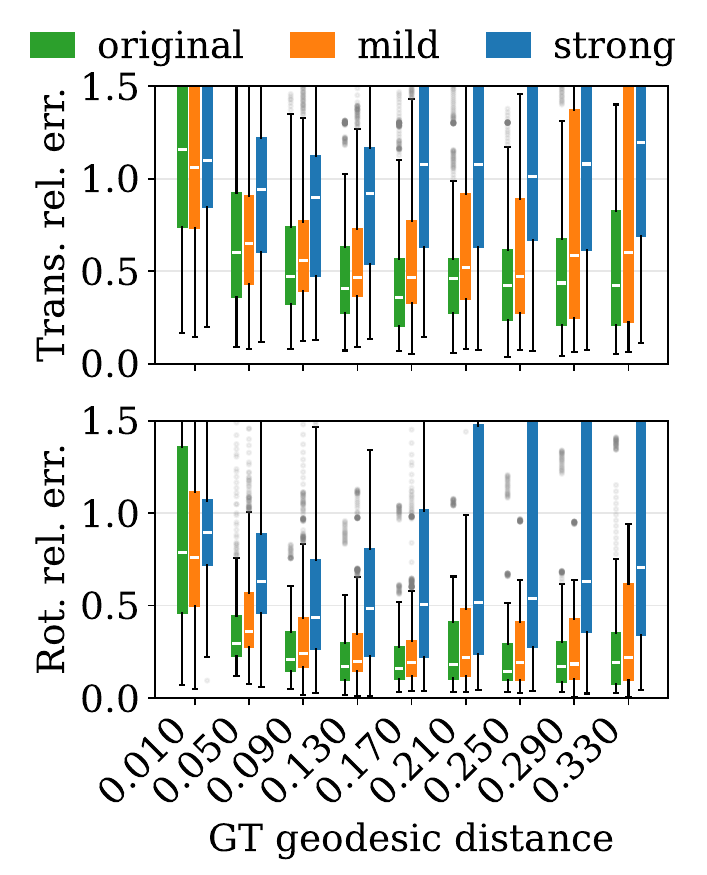}
\end{minipage}
\begin{minipage}[c]{0.145\columnwidth}
\centering
\includegraphics[width=\linewidth]{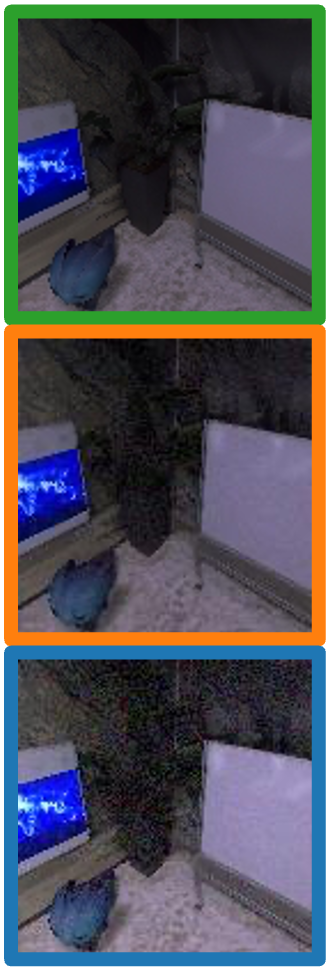}
\end{minipage}
\caption{\textbf{(Left)} Amount of error change after convergence, given the initial error amount. \textbf{(Right)} The error trends of the proposed method under different amounts of photometric noise. The photometric noise mimics a short exposure time, which is realistic for the proposed experimental setup.}
\label{fig:basin_and_noise}
\vspace{-0.15cm}
\end{figure}

\section{Experiments}
\subsection{Realistic Photometric Prior} 

\subsubsection{PPA-like Photometric Noise}
To emulate short-exposure photometric degradation expected in PPAs, we use a Poisson--Gaussian camera model \cite{4623175}. Example images of the Replica~\cite{straub2019replica} dataset degraded with this model are shown at the rightmost side of~\cref{fig:basin_and_noise}. The center graph shows that \textit{mild} noise reduces the convergence basin, while \textit{strong} noise breaks our method. The following paragraph provides a more detailed explanation of each parameter of the degradation model.

For a clean linear pixel intensity $x\in[0,1]$, we first apply Photo-Response Non-Uniformity (PRNU):
\begin{equation}
x'=\mathrm{clip}\!\left(x(1+\eta),0,1\right),\quad \eta\sim\mathcal{N}(0,\sigma_p^2).
\end{equation}
Let $s=\alpha F x'$ be the expected signal electrons, where $\alpha$ is the exposure scale and $F$ is full-well capacity. The noisy electron count is
\begin{equation}
n_e \sim \mathrm{Poisson}(s)+\mathrm{Poisson}(\lambda_d t_e)+\mathcal{N}(0,\sigma_r^2),
\end{equation}
with dark-current rate $\lambda_d$ (e$^-$/s), exposure time $t_e$ (s), and read-noise standard deviation $\sigma_r$ (e$^-$). The output is quantized by gain $g$ (e$^-$/DN) and ADC bit depth $b$:
\begin{equation}
n_{\mathrm{DN}}=\frac{\mathrm{clip}(n_e/g,\,0,\,2^b-1)}{2^b-1}.
\end{equation}

By denoting the parameter vector as $\boldsymbol{\theta}=(\alpha,\sigma_r,\lambda_d,\sigma_p,F,g,b,t_e)$, the two settings are as follows:
\begin{equation}
\begin{aligned}
\boldsymbol{\theta}_{\text{strong}}=&(0.20,\;25,\;1.0,\;0.01,\;15000,\;1.5,\;10,\;0.001),\\
\boldsymbol{\theta}_{\text{mild}}=&(0.12,\;8,\;0.3,\;0.003,\;15000,\;1.5,\;10,\;0.001).
\end{aligned}
\end{equation}

Thus, compared with \emph{mild}, the \emph{strong} option uses larger $\alpha$, $\sigma_r$, $\lambda_d$, and $\sigma_p$, yielding stronger heteroscedastic noise and earlier highlight saturation, while keeping the same sensor/ADC baseline $(F,g,b,t_e)$.

\subsubsection{Real-world Sequences}\label{sec:realworld}
Evaluating on more realistic premises may be beneficial to measure feasibility gap of the method; however, the existing widely-used \textit{real-world RGB datasets are mostly out of our extremely-small frame-to-frame baseline assumption of PPAs}~\cite{bose2025descriptor}.

Despite this unfavorable setting, we evaluate on the popular VO/SLAM dataset TUM, as \textit{PPA images cannot currently be streamed off-sensor at their native capture framerate}, which is also why our main experiments are conducted on Replica.
We report mean RPE of frame pairs with 10-frame gap, of our method with DSINE estimated normal and two baselines. For fair comparison, the hyperparameters of baselines are finetuned on these sequences, and DROID-SLAM (DSLAM) ran frontend-only. DROID-SLAM ran on full resolution was reported as a reference with commodity camera. Pairs where all methods successfully gave valid estimation on are considered. With 128 by 128 resolution, RPE is not much worse than DROID-SLAM and success rate is higher than DSO, which struggles with low resolution.

\begin{table}[h]
\centering
\setlength{\tabcolsep}{4pt}
\caption{Comparison with VOs on real-world sequences.}
\resizebox{0.75\linewidth}{!}{\begin{tabular}{ll|c c c | c}  
\toprule
    & TUM & Ours+DSINE & DSO & DSLAM & DSLAM (full-res) \\
    \midrule
    \multirow{3}{*}{\rotatebox[origin=c]{90}{ fr2\_xyz}}& RPE-r ($^\circ$) & 4.70 & 0.83 & 2.36 & 1.20\\
    & RPE-t (mm) & 20.3 & 7.6 & 16.5 & 11.1\\
    & Success rate (\%) & 61 & 45 & 100 & 100\\  
\bottomrule
\end{tabular}}
\end{table}

\subsection{Nonlinearities and Failure Modes} 
The nonlinearity of $\mathbb{SE}(3)$ renders our objective nonlinear as well; consequently, convergence of the GBP iterations cannot be guaranteed for arbitrary camera-pose initializations. To examine the convergence basin with respect to camera pose initialization, we constructed the leftmost plot in~\cref{fig:basin_and_noise} using all image pairs from the evaluation set, where the x-axis denotes the initial pose error and the y-axis denotes the change in error after optimization. Rotation generally converges when the initialization is sufficiently close to the ground truth, whereas translation tends to converge more reliably under moderately larger motion, as extremely small disparities make depth discontinuities difficult to resolve.

Note that the camera pose errors reported here are angular pose errors for both translation and rotation. This is a standard metric in image matching~\cite{sarlin2020superglue} and relative camera pose estimation~\cite{jin2021image}, where the primary concern is an image pair rather than temporal error accumulation over a sequence.

\subsection{Pose Accuracy Compared with Centralized Visual Odometries}
In this section, our method is compared against other centralized direct odometry baselines may further show the practicality gap, and we provide this for the same Replica frame pairs on the TUM protocol given at~\cref{sec:realworld}.

\begin{table}[h]
\centering
\setlength{\tabcolsep}{4pt}
\caption{Comparison against centralized VOs.}
\resizebox{0.8\linewidth}{!}{\begin{tabular}{l|c c c c | c}  
\toprule
    Replica & Ours+GT & Ours+DSINE & DSO & DSLAM & DSLAM (full-res) \\
    \midrule
    RPE-r ($^\circ$) & 7.29 & 7.21 & 5.62 & 5.64 & 2.34\\
    RPE-t (mm) & 91.3 & 99.3 & 70.4 & 85.3 & 34.1\\
    Success rate (\%) & 97 & 93 & 78 & 100 & 100\\
\bottomrule
\end{tabular}}
\end{table}

\subsection{Comparison with SoTA Geometry-guided VO}
In the left table of \cref{tab:VO}, we compare our method against state-of-the-art optimization-based visual odometry approaches that leverage network-inferred monocular geometric priors. For the definition of the Pose Error AUC metric used here, please refer to~\cite{dai2017scannet}.

As we were unable to identify a state-of-the-art VO method that uses a normal prior, we instead compare against MADPose~\cite{yu2025relative}, which employs a monocular depth prior. As shown in the table, although the proposed method yields substantially lower camera pose accuracy than MADPose when the latter is combined with a modern, computationally intensive matching pipeline (SuperPoint~\cite{detone2018superpoint} detection with LightGlue~\cite{lindenberger2023lightglue} matching), it achieves camera pose accuracy comparable to MADPose paired with a classical matching pipeline (ORB detection with Nearest Neighbor matching) while requiring only about one-fifth of the computational cost.

Furthermore, obtaining the monocular depth prior used in MADPose requires DepthAnythingV2~\cite{yang2024depth}, a Transformer-based model. The all-to-all attention mechanism in such models has not, to our knowledge, been successfully implemented on pixel-processor arrays; moreover, even if such an implementation were possible, it would introduce a severe inter-pixel communication bottleneck. This constitutes an additional practical challenge that is not reflected in the FLOP counts reported in our table.

\subsection{Scalability}
The proposed pixel-distributed optimization framework has the advantage that the computational workload assigned to each pixel is highly homogeneous. As a result, the method can be applied directly to cameras with different resolutions without any architectural modification. In this section, we leverage this property to evaluate our approach across multiple image resolutions and analyze the resulting performance.

\begin{figure}[t]
    \centering
    \includegraphics[width=1\linewidth]{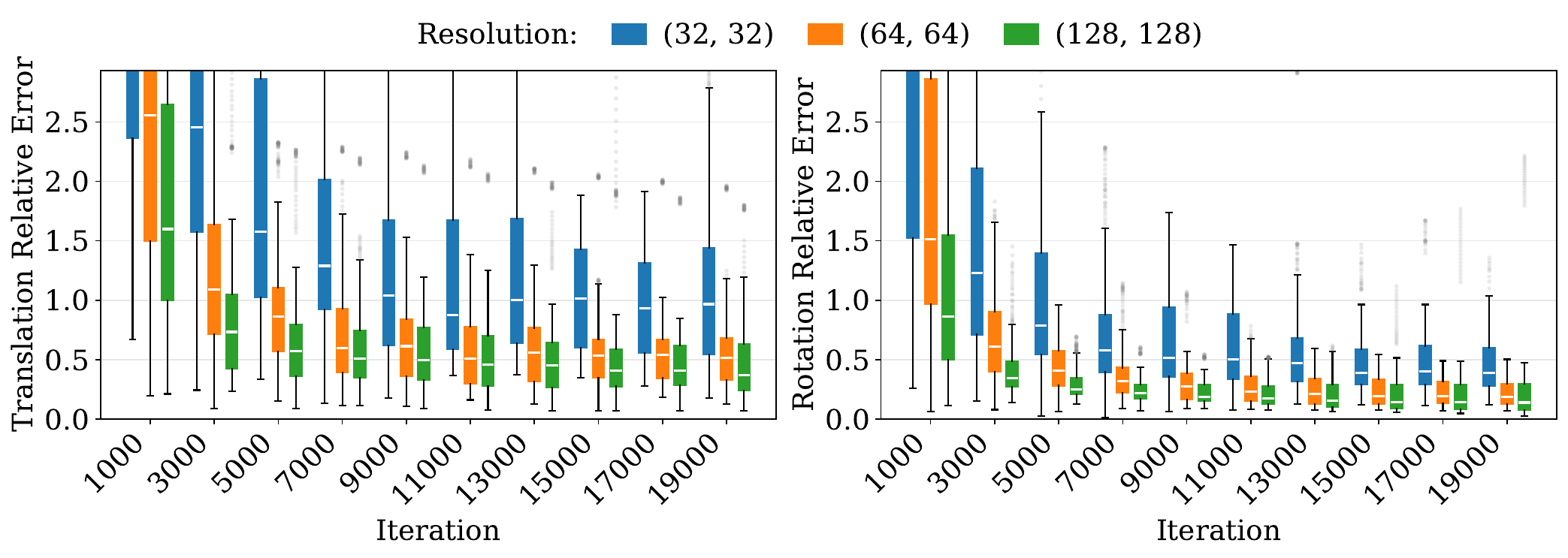}
    \caption{Error of estimated poses across various resolutions of input images.}
    \label{fig:resolution}
    \vspace{-0.3cm}
\end{figure}

As shown in \cref{fig:resolution}, when the resolution is reduced by half, the system still achieves performance comparable to that of the original resolution. In contrast, when the resolution is reduced to one quarter, the optimization fails to converge to the correct solution. This degradation arises from the fact that, at such low resolutions, pixel correspondences cannot be established with sufficient fine-grained detail to support a reliable camera pose estimation. This behavior is intuitive, as coarse spatial sampling inherently limits the precision of correspondence-based inference.

\begin{figure}[t]
    \centering
    \includegraphics[width=1\linewidth]{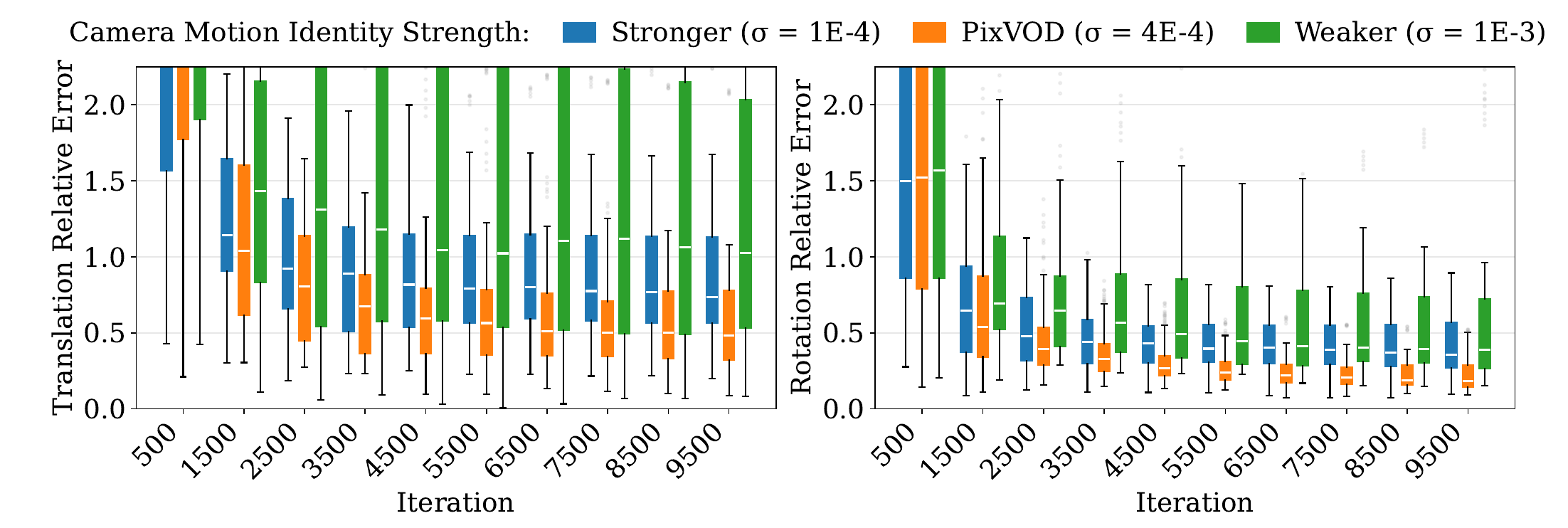}
    \caption{Error of estimated poses across various strengths of the camera motion identity factor.} 
    \label{fig:regularization_ablation}
    \vspace{-0.3cm}
\end{figure}
\begin{figure}[t]
    \centering
    \includegraphics[width=1\linewidth]{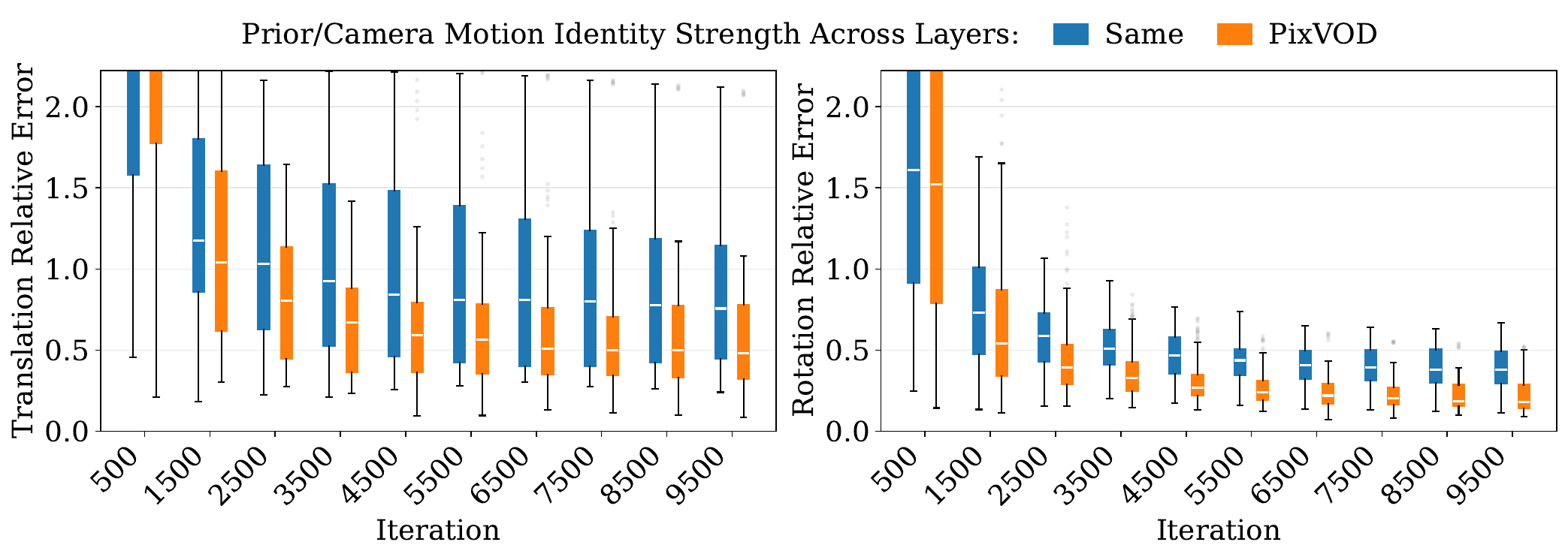}
    \caption{Error of estimated poses across various strategies of assigning camera pose identity and prior strengths on different layers of sharded factor graph topology.}
    \label{fig:pyramid_ablation}
    \vspace{-0.3cm}
\end{figure}
\begin{figure}[t]
    \centering
    \includegraphics[width=1\linewidth]{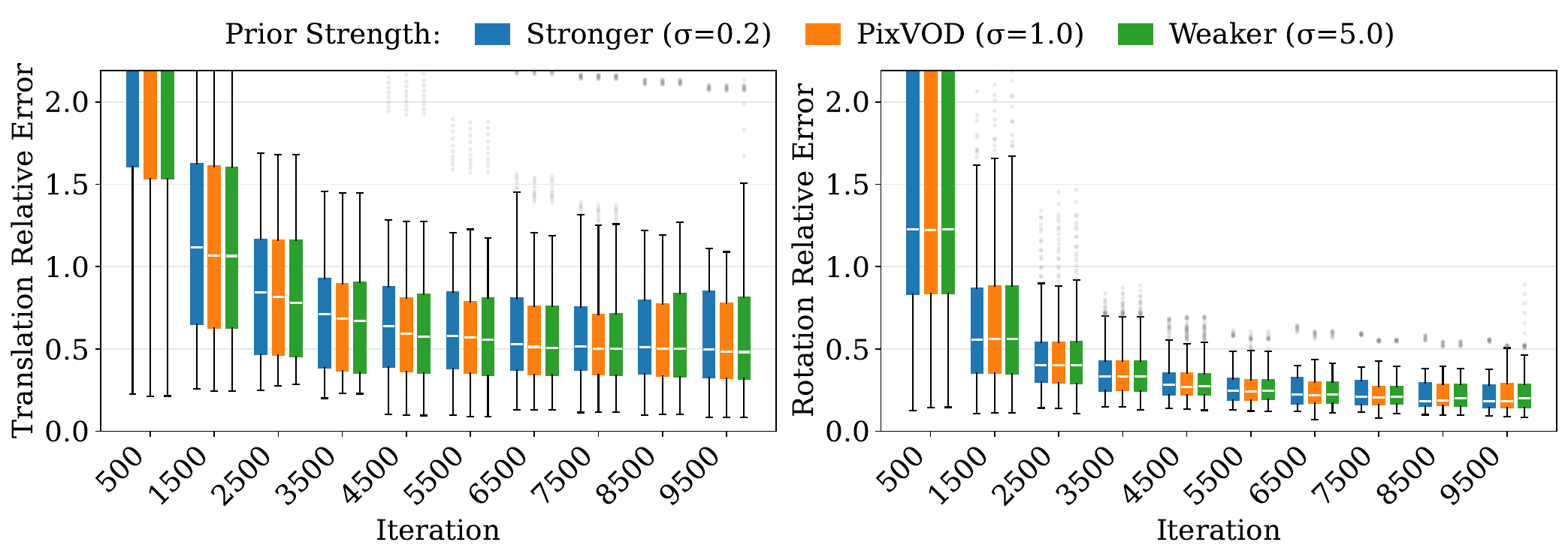}
    \caption{Error of estimated poses across various strengths of the prior factor.}
    \label{fig:prior_ablation}
    \vspace{-0.3cm}
\end{figure}
\begin{figure}[t]
    \centering
    \includegraphics[width=1\linewidth]{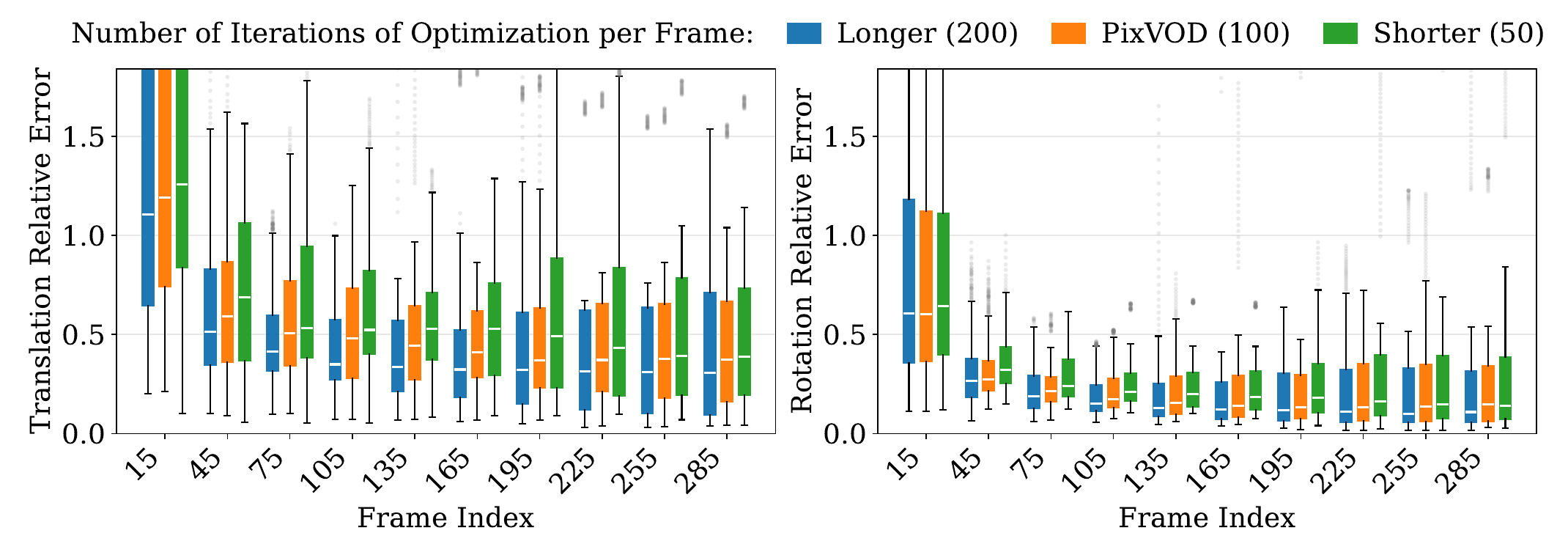}
    \caption{Error of estimated poses across various numbers of iterations to optimize per frame.}
    \label{fig:iter_optim_ablation}
    \vspace{-0.3cm}
\end{figure}

\section{Ablation Studies}
This section provides analysis and insights into the hyperparameter settings that were adopted without extensive validation in Sec.~4.2 of the main paper, due to space constraints.

\subsection{Camera Motion Identity Strength}
This section analyzes how the strength of the camera motion identity factor affects convergence, similar to~\cite{alzugaray2024pixro}. As shown in \cref{fig:regularization_ablation}, an overly weak camera motion identity factor fails to enforce effective consensus across pixels, causing the optimization to converge to inaccurate camera pose estimates. Conversely, when the camera motion identity factor is too strong, the photometric factors—which are the primary source of informative constraints in our system—are effectively downweighted, and the optimization again converges to inaccurate camera poses.


\subsection{Inter-layer Difference of Regularization}
This section experimentally validates the rationale for increasing the strength of the camera motion identity factor and the prior factor in higher layers (by a factor of $0.5$ per layer) of the hierarchy. As shown in \cref{fig:pyramid_ablation}, assigning identical factor strengths to all layers not only slows convergence but also leads the optimizer to settle at a more suboptimal solution. (For identical factors, the strengths are set to be the geometric mean of the factor strengths of all layers of the original strategy, which are $\sigma_R=3.536\times10^{-5}$ and $\sigma_P=8.839\times10^{-2}$.) This behavior aligns with the intuition discussed in Sec.~4.2 regarding the differing reliability of pose estimates between lower and higher levels. Note that the inverted strategy, when the higher layers of the topology have weaker factor strengths than lower layers, is not shown on the graph, as it is numerically unstable and often diverges. 

\subsection{Prior Strength}
As shown in~\cref{fig:prior_ablation}, the camera pose estimation of our system is generally robust to variations in the strength of the prior-factor. However, excessively large priors tend to drive the solution toward incorrect poses since they attract the solution toward zero (Note the prior is set to zero to act similarly as $L_2$ regularizer for optimization stability), whereas overly small priors---though not apparent in the accompanying box-plot visualization---often introduce numerical instability that leads to divergence in the early stages of optimization. Consequently, we select the weakest prior that avoids such early-stage divergence.

\subsection{Number of Iterations per Frame}
As shown in \cref{fig:iter_optim_ablation}, increasing the number of optimization iterations per frame generally improves the accuracy of camera pose estimation, which is consistent with intuition. The improvement is particularly pronounced for translation. However, as indicated by the box plots for the post-convergence region (after 105-th frames' bins) of the translation graph, once a sufficient number of iterations has been performed within each frame, further excessive optimization yields only marginal accuracy gains in most cases. This characteristic is advantageous for PPA, since external factors, such as illumination changes or rapid camera motion, may require adaptive adjustment of the frame acquisition rate, which in turn implies that the number of feasible optimization iterations per frame may also vary.

\end{document}